\documentclass[a4paper]{article}%
\usepackage{placeins}
\usepackage{amssymb}
\usepackage{graphicx}
\usepackage{rotating}
\usepackage{amsfonts}
\usepackage{mathrsfs}
\usepackage{amsmath}
\usepackage{babel}
\usepackage{lastpage}
\usepackage[utf8]{inputenc}
\usepackage[legalpaper,bookmarks=true,colorlinks=true,linkcolor=blue,citecolor=blue]{hyperref}
\usepackage{graphicx}%
\usepackage{fancyhdr,enumerate}
\usepackage{color}
\usepackage[mathlines]{lineno}
\usepackage{lscape}
\usepackage{subfigure}
\usepackage{epsfig}
\usepackage[sort&compress]{natbib}
\usepackage{epstopdf}
\usepackage{subfigure}
\usepackage{float,adjustbox}
\usepackage{bm}
\usepackage{booktabs}
\usepackage{geometry}
\usepackage{geometry}
\begin{document}

\title{Prediction of Coffee Ratings Based On Influential Attributes Using SelectKBest and Optimal Hyperparameters}

\author{Edmund F. Agyemang$^{\textbf{1,2,3*}}$, Lawrence Agbota$^{2}$, Vincent Agbenyeavu$^{2}$, \\ Peggy Akabuah$^{2}$, Bismark Bimpong$^{2}$ \& Christopher Attafuah$^{2}$\\
$^{\textbf{1}}$Department of Biostatistics and Data Science, Celia Scott Weatherhead School of \\ Public Health and Tropical Medicine, Tulane University, New Orleans, 
Louisiana, USA\\	
$^{\textbf{2}}$	School of Mathematical and Statistical Science, College of Sciences,\\ University of Texas Rio Grande Valley, Edinburg, Texas, USA.\\
$^\textbf{{3}}$Department of Computer Science \& Information Systems, Ashesi University, Ghana.\\
Corresponding author: \color{blue}edmundfosu6@gmail.com, ORCID ID: 0000-0001-8124-4493}

\date{}
\maketitle
\begin{abstract}
	\noindent	
This study explores the application of supervised machine learning algorithms to predict coffee ratings based on a combination of influential textual and numerical attributes extracted from user reviews. Through careful data preprocessing including text cleaning, feature extraction using TF-IDF, and selection with SelectKBest, the study identifies key factors contributing to coffee quality assessments. Six models (Decision Tree, K-Nearest Neighbors, Multi-layer Perceptron, Random Forest, Extra Trees, and XGBoost) were trained and evaluated using optimized hyperparameters. Model performance was assessed primarily using F1-score, G-mean, and AUC metrics. Results demonstrate that ensemble methods (Extra Trees, Random Forest, and XGBoost), as well as Multi-layer Perceptron, consistently outperform simpler classifiers (Decision Trees and K-Nearest Neighbors) in terms of evaluation metrics such as F1 scores, G-mean and AUC. The findings highlight the essence of rigorous feature selection and hyperparameter tuning in building robust predictive systems for sensory product evaluation, offering a data driven approach to complement traditional coffee cupping by expertise of trained professionals.\\
	
\end{abstract}


\newpage
\section{Introduction}	
Coffee, often hailed as the world’s favorite beverage, holds a unique place in both global agriculture and daily life. Its origins trace back to the ancient coffee forests of Ethiopia, where history has it that a goat herder named Kaldi first discovered the energizing effects of the coffee bean \cite{pendergrast2010uncommon}. Since then, coffee cultivation and consumption have spread across continents, shaping economies, cultures, and even social rituals. Today, coffee is one of the most valuable agricultural commodities worldwide, second only to crude oil in dollar value among globally traded goods \cite{pancsira2022international}. The coffee industry provides livelihoods for an estimated 125 million people, many of whom are smallholder farmers in developing countries across Africa, Latin America, and Asia. Major coffee-producing nations include Brazil, Vietnam, Colombia, Ethiopia, and Honduras, with Brazil consistently leading global output \cite{ICO2025CoffeeMarket}. The consumption of coffee has evolved alongside its production. Once a luxury for the elite, coffee has become an everyday staple, with more than two billion cups consumed each day across the globe \cite{jhaveri2021coffee}. Its popularity is deeply intertwined with social and cultural practices, from the European cafe culture of the Enlightenment era to the modern specialty coffee movement that emphasizes origin, quality, and sustainability. Given coffee’s immense economic, social, and cultural significance, understanding the attributes that influence its ratings remains an important focus for researchers, producers, and consumers alike.\\

\noindent
Traditionally, assessing the quality of products like coffee, wine, or tea  has relied heavily on the expertise of trained professionals or professional human expertise, known as “cuppers." These experts evaluate coffee samples through a standardized process called cupping, which involves smelling, tasting, and scoring brewed coffee to judge its aroma, flavor, acidity, body, and aftertaste \cite{sca2018cupping}. The cupping procedure follows strict protocols to minimize bias, including the use of specific ratios, temperatures, and blind tasting techniques. Despite its widespread use and importance in the industry, the traditional approach has some limitations. Human assessment is inherently subjective, inconsistent and can be influenced by personal preferences, fatigue, or even environmental factors. Additionally, traditional methods can be time consuming and require significant training to ensure consistency and reliability. Nevertheless, sensory evaluation by skilled cuppers remains the gold standard for determining coffee quality and guiding both producers and buyers.\\

\noindent
 Machine learning algorithms, on the other hand, are able to process large amounts of data ranging from chemical composition and physical attributes to sensory scores and identify patterns that may not be immediately obvious to human evaluators. In recent years, machine learning has become a transformative tool for quality prediction within the food and beverage industry. In the coffee sector, researchers have used machine learning techniques to classify coffee beans by sustainability labeling, organic, fair trade, country of origin, type of coffee (espresso, filter, instant, single cup, with milk, iced) and intrinsic attributes (roast degree, specialty coffee), and extrinsic attributes (brands and packaging) \cite{samoggia2018coffee}.
The prediction of coffee ratings based on influential attributes and optimal hyperparameters through supervised machine learning algorithms represents a significant advancement in the coffee industry. By employing various machine learning techniques, researchers aim to automate the assessment of coffee quality, providing a reliable framework for predicting ratings based on key attributes like aroma, flavor, acidity, body, balance, and aftertaste. This integration of data science enhances the understanding of coffee characteristics, which is crucial for both consumers and producers in guiding purchasing decisions and improving product standards.\\

\noindent
\cite{barahona2020sensory} research in this domain has highlighted the complexities involved in evaluating coffee quality, encompassing challenges such as the variability in study designs, inconsistent definitions of coffee quality, and the need for comprehensive datasets. These challenges can hinder the development of robust predictive models and may lead to overfitting or bias in algorithm performance, necessitating a careful approach to data collection and preprocessing \cite{liu2023establishing}.
Furthermore, machine learning offers promising solutions to address these issues, as evidenced by successful applications that have achieved high classification accuracies, demonstrating the potential of technology to reshape traditional methods of quality assessment. Despite its advancements, the field is not without controversies, as debates persist regarding the definition and measurement of coffee quality, as well as the accessibility of machine learning tools to stakeholders beyond experts \cite{liu2023establishing}.\\

\noindent
In this study, we apply supervised machine learning techniques to build  predictive models. The data consists of user's rating of coffee and other relevant attributes. six (6) distinct algorithms namely Decision Trees (DT),  K-Nearest Neighbors (KNN), Multi-layer Perceptrons (MLP), Random Forest (RF), Extra Trees (Extremely Randomized Trees) and Extreme Gradient Boosting (XGBoost) are utilized to generate the outputs (predicted ratings) from the attributes that are used as inputs to the models. The model performance is evaluated against the actual ratings and the F1-score, G-mean and AUC are the main metric used as assessment. The remainder of the paper is organized as follows:  section \ref{sec2} discusses the data preprocessing via text data including feature extraction and selection. In section \ref{sec3}, the  machine learning algorithms adopted for the study and implementation are elaborated on. Section \ref{sec4} delves into the results, findings and discussions of the study encompassing any insightful findings of the relationships between coffee attributes and ratings and key challenges encountered during execution of the project. Lastly, section \ref{sec5} concludes the study and provide recommendations for further work.

\section{Data Preprocessing \label{sec2}}
The data preprocessing step is pivotal for converting unrefined data into a state that can be effectively employed by machine learning algorithms. For this project, the dataset consists of both numerical and textual features. We focus on cleaning and transforming these features into usable formats, especially for the textual review data.	
\subsection{Text Data Preprocessing}
The textual reviews, which are unstructured, require thorough preprocessing to extract meaningful features. The following steps were applied and the resulting cleaned data is a set of processed text reviews that are ready to be transformed into numerical features:
\begin{itemize}
\item [$\star$] \textbf{Cleaning Text}: First, we removed non-alphabetical characters using regular expressions, keeping only words made up of letters.
\item [$\star$] \textbf{Lowercasing}: All text was converted to lowercase to maintain uniformity and avoid redundancy.
\item [$\star$] \textbf{Tokenization and Lemmatization}: Each review was split into words (tokens), and these tokens were lemmatized using WordNetLemmatizer from the NLTK library. Lemmatization ensures that words like “running" and “ran" are reduced to their root form, “run."
\item [$\star$] \textbf{Stopword Removal}: Common stopwords such as “and," “the," “is," etc., were removed from the reviews using NLTK’s list of English stopwords. These words do not carry significant meaning and could add noise to the analysis.
\end{itemize}

\subsection{Feature Extraction: Text to Vector}
After preprocessing the text data, reviews were transformed  into numerical representations that machine learning models can understand.  The \texttt{TfidfVectorizer} from scikit-learn was used to convert the text data into a matrix of TF-IDF features. TF-IDF is a statistical measure used to evaluate the importance of a word in a document relative to a collection of documents \cite{park2025automated}. This method gives greater weight to words that are common in a single review but not across all the reviews, thereby making the informative words more prominent \cite{danyal2024sentiment}. The \texttt{TfidfVectorizer} was fit on the training data, then used to transform both the training and test datasets using the features that were learned from the training data. The same \texttt{TfidfVectorizer} was applied to the test data to ensure the transformed test set maintains the same feature space as the training set. The result of this transformation is a matrix of numerical feature vectors for each review. These vectors, which represent the importance of words in the reviews, are then used as inputs for machine learning models.

\subsection{Feature Selection}
For efficient model training, irrelevant or less informative features were filtered out. This was done using:
\begin{itemize}
	\item [$\star$] \textbf{Variance Thresholding}: A feature selection method that removes features with low variance across samples.
	\item [$\star$] \textbf{SelectKBest}: This method was used to select the most informative features by using statistical tests (e.g., ANOVA F-test). For this project, we selected a subset of features (10) that contributed the most to distinguishing between coffee ratings. 
\end{itemize}
The data was now ready for building the three (3) models namely DT, KNN and MLP, with the textual data transformed into a usable numerical format and irrelevant features removed.

\newpage
\section{Algorithms and Implementation \label{sec3}}
The six (6) methods adopted for the study and how they were implemented are discussed below:  

\subsection{Decision Tree (DT)}
The Decision Tree classifier was implemented using \texttt{DecisionTreeClassifier} from scikit-learn. Decision trees create a model by splitting the data based on the most significant features, which is particularly useful for both categorical and numerical data \cite{james2023tree}. For this project, hyperparameter tuning was performed using GridSearchCV, optimizing parameters like \texttt{max\_depth}, \texttt{min\_samples\_split}, and \texttt{criterion}. The tree was trained using \texttt{gini} and \texttt{entropy} criterion, which measures the impurity of a split. Important hyperparameters such as \texttt{max\_depth} and \texttt{min\_samples\_leaf} were tuned to prevent overfitting \cite{agbota2024enhancing}.

\subsection{K-Nearest Neighbors (KNN)}
KNN classifies a data point based on the majority class of its nearest neighbors \cite{kumar2025improving,agyemang2024modelling,agyemang2025addressing}. In this project, KNN was implemented using \texttt{KNeighborsClassifier} from scikit-learn. Hyperparameter tuning was carried out using GridSearchCV, optimizing parameters such as the number of neighbors (\texttt{n\_neighbors}) and the distance metric (\texttt{metric}). The model used the \texttt{euclidean} distance metric for calculating the proximity between data points. The number of neighbors (\texttt{n\_neighbors}) and the weight function (\texttt{weights}) were optimized using cross-validation.

\subsection{Multi-layer Perceptron (MLP)}
The Multi-layer Perceptron (MLP) is a type of feedforward neural network consisting of an input layer, one or more hidden layers, and an output layer \cite{mehrkash2025robustness}. For this project, the MLP model was implemented using the \texttt{MLPClassifier} from scikit-learn. The model was trained using the transformed TF-IDF features from the text data. Hyperparameter tuning was performed through GridSearchCV, optimizing parameters such as the number of hidden layers, activation function, and solver.

\subsection{Random Forest (RF)}
RF has the ability to reduce overfitting by averaging multiple decision trees trained on different subsets of the data \cite{halabaku2024overfitting,sakyi2025heart}. In this study, RF was implemented by first selecting the top $k$ features for $k=10,15,20,25$ using \texttt{SelectKBest} with the \texttt{f\_classif} score function, based on training TF-IDF data. It then splits the data into training and validation sets using \texttt{train\_test\_split}. Hyperparameter tuning is performed using \texttt{GridSearchCV} to optimize the Random Forest model's parameters, including the number of estimators (\texttt{n\_estimators}), split criterion (\texttt{criterion}), maximum tree depth (\texttt{max\_depth}), minimum samples required to split or be at a leaf node, and the number of features considered at each split (\texttt{max\_features}). The best hyperparameters are identified, and the final Random Forest model is trained with these optimal parameters. 

\subsection{Extra Trees (Extremely Randomized Trees)}
Similar to Random Forest but with additional randomness in tree splitting \cite{bicego2023good,sakyi2025heart}. The model begins by selecting the top $k$ features for $k=10,15,20,25$ from the training TF-IDF data using \texttt{SelectKBest} with the \texttt{f\_classif} scoring function. Afterward, the data is split into training and validation sets using \texttt{train\_test\_split}. Hyperparameter optimization is done with \texttt{GridSearchCV}, which searches over various combinations of parameters such as the number of estimators (\texttt{n\_estimators}), the criterion used for splitting (\texttt{criterion}), the maximum depth of trees (\texttt{max\_depth}), the minimum number of samples required to split or be at a leaf node (\texttt{min\_samples\_split} and \texttt{min\_samples\_leaf}), and the maximum number of features considered at each split (\texttt{max\_features}). The best hyperparameters are chosen based on 5-fold cross-validation and the F1-weighted score. Using the best-found hyperparameters, the final model is trained on the training data. 

\subsection{Extreme Gradient Boosting (XGBoost)}
XGBoost is known for its high performance and efficiency \cite{hakkal2024xgboost,sakyi2025heart}. The implementation of XGBoost is similar to that of the Random Forest but here using these optimal hyperparameters, the model is trained with the \texttt{XGBClassifier}.

\newpage
\subsection{Model Performance Metrics}	
All six (6) models were evaluated mainly using the G-mean, AUC and F1 score, a type of precision-recall metric used in many classification settings, particularly where the datasets have class imbalance \cite{sathyanarayanan2024confusion}. The F1 score was computed on both the training and validation sets to assess the model's ability to generalize. Cross-validation was used to estimate the model's performance and avoid overfitting. For each model, 5-fold cross-validation was applied, where the data is split into 5 subsets, and the model is trained and validated 5 times, each time using a different subset as the validation set. The F1 scores from each fold were averaged to provide a reliable performance measure.\\

\noindent
The computational formulas for weighted precision, recall and F1 score are given respectively by (\ref{A22}), (\ref{A33}) and (\ref{A44}): 

\begin{equation}
	\text{Weighted Precision} = \frac{\sum_{j=1}^n w_j \times (\text{True Positives in class} j)}{\sum_{j=1}^n w_j \times (\text{Predicted Positives in class } j)}
	\label{A22}
\end{equation}

\begin{equation}
	\text{Weighted Recall} = \frac{\sum_{j=1}^n w_j \times (\text{True Positives in class} j)}{\sum_{i=j}^n w_j \times (\text{Actual Positives in class} j)}
	\label{A33}
\end{equation}

\begin{equation}
	\text{Weighted F1 Score} = 2 \times \frac{\text{Weighted Precision} \times \text{Weighted Recall}}{\text{Weighted Precision} + \text{Weighted Recall}}	
	\label{A44}
\end{equation}
Additionally, the study employed the G-mean statistic given in (\ref{q4}),
which is also an excellent indicator that effectively handles imbalanced class issues.
\begin{equation}
	G-mean=\sqrt{\text{(Recall/Sensitivity)} \times \text{Specificity}}	
	\label{q4}
\end{equation}
The G-mean is the geometric mean value used to measure overall model performance. Poor classification results will produce a small G-mean value \citep{bekkar2013evaluation}.\\

\begin{figure}[!hbt]
	\centering
	\includegraphics[width=0.8\linewidth]{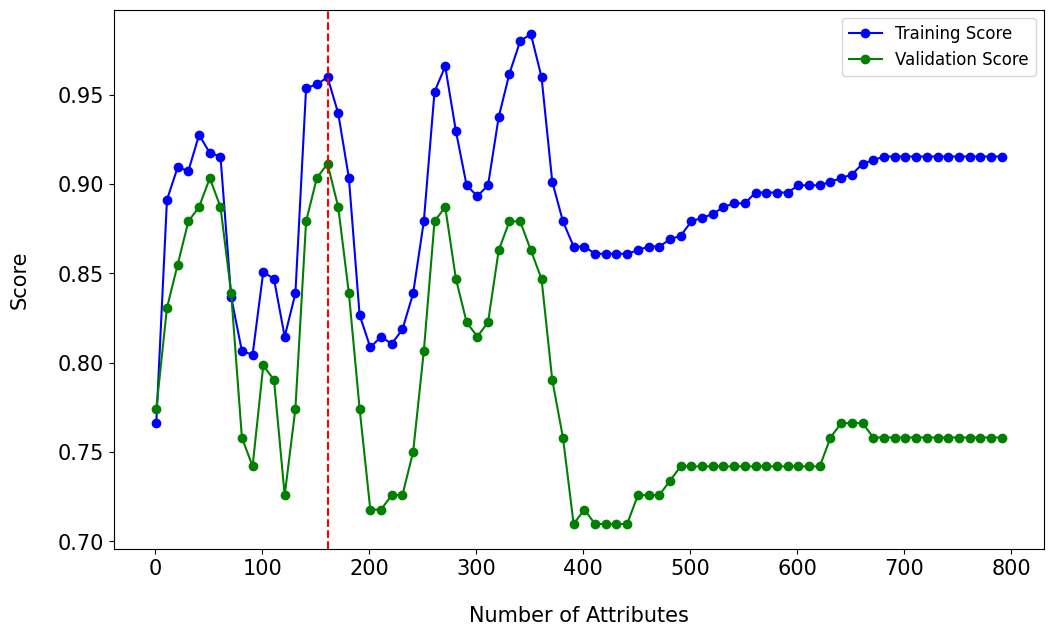}
	\caption{\textbf{Training and Validation Scores versus Attributes}}
	\label{AA}
\end{figure}

\noindent
 Figure \ref{AA}, presents the graphical display of the training and validation scores versus the number of optimal attributes.

\noindent
From Figure \ref{AA}, it is observed that $161$ optimal attributes were chosen (shown by the straight red line in Figure \ref{AA}) using the training and validation scores by setting a variance threshold for the maximum acceptable difference to $5\%$.

\newpage
\section{Results, Findings and Discussion \label{sec4}}	
This section provides a detailed discussions of the findings of the study.

\subsection{Insightful Findings for $k=10$}
Rigorous hyparameter tuning via GridSearchCV resulted in the following optimal hyperparameters  for the six (6) models for $k=10$. Decision Tree: criterion=\texttt{gini}, max\_depth=3, max\_features=161, min\_samples\_leaf=1, min\_samples\_split=5.  Random Forest: criterion=\texttt{gini}, max\_depth=10, max\_features=\texttt{sqrt}, min\_samples\_leaf=2, min\_samples\_split=10, n\_estimators=50.  Extra Trees: criterion=\texttt{entropy}, max\_depth=None, max\_features=\texttt{sqrt}, min\_samples\_leaf=2, min\_samples\_split=5, n\_estimators=200.  XGBoost: colsample\_bytree=1.0, gamma=1, learning\_rate=0.1, max\_depth=5, n\_estimators=100, subsample=0.6. K-Nearest Neighbors: metric=\texttt{euclidean}, n\_neighbors=7, weights=\texttt{uniform}.   
Multi-Layer Perceptron: activation=\texttt{relu}, alpha=0.0001, hidden\_layer\_sizes=(150,), learning\_rate=\texttt{constant}, solver=\texttt{adam}.

\begin{table}[!htbp]
	\centering
	\caption{Model Performance Evaluation of DT, KNN, MLP, NB, RF and XGBoost for $k=10$}
	\begin{tabular}{lccccccccc}
		\toprule
		\textbf{Models}	& &  \multicolumn{2}{c}{\textbf{Decision Tree}} && \multicolumn{2}{c}{\textbf{K-Nearest Neighbors}}&&\multicolumn{2}{c}{\textbf{Multi-layer Perceptron}}\\ 
		
		\cmidrule{3-4} \cmidrule{6-7}\cmidrule{9-10}  \vspace{0.05in}
		&& Training   & Validation &&Training  & Validation  &&Training&Validation\\ 
		\hline
		\textbf{Recall (TPR)} 	&  & 0.8810 & 0.8306 & & 0.9073 &0.8065 &&0.9113&0.8629 \\
		\textbf{Specificity (TNR)} &  & 0.7101 & 0.6250 & &  0.8047 &0.6750 &&0.8166&0.7500 \\ 
		\textbf{Precision} 	&  & 0.8855 & 0.8283 & & 0.9076 &0.8042 &&0.9114&0.8611\\ 
		\textbf{F1-Scores} 	&  & 0.8766 & 0.8242 & & 0.9057 & 0.8051 &&0.9100&0.8614 \\ 
		\textbf{G-Mean} &  & 0.7909 & 0.7205 & & 0.8545 &0.7378 &&0.8626&0.8045 \\   
		\textbf{AUC} &  & 0.8477 & 0.7815 & & 0.9581 &0.8810 &&0.9500&0.9426 \\ 
		\bottomrule
\textbf{Models}	& &  \multicolumn{2}{c}{\textbf{Extra Trees}} && \multicolumn{2}{c}{\textbf{Random Forest}}&&\multicolumn{2}{c}{\textbf{XGBoost}}\\ 

\cmidrule{3-4} \cmidrule{6-7}\cmidrule{9-10}  \vspace{0.05in}
&& Training   & Validation &&Training  & Validation  &&Training&Validation\\ 
\hline
\textbf{Recall (TPR)} 	&  & 0.9214& 0.8710 & & 0.9214& 0.8710 &&0.9214&0.8468 \\
\textbf{Specificity (TNR)} &  & 0.8284 & 0.7250 & & 0.8343 &0.7750 &&0.8521&0.7500 \\ 
\textbf{Precision} 	&  & 0.9220 & 0.8698 & &  0.9217 &0.8697 &&0.9210&0.8459 \\ 
\textbf{F1-Scores} &  & 0.9201 & 0.8680 & & 0.9203 &0.8701 &&0.9207&0.8463 \\ 
\textbf{G-Mean} &  & 0.8737 & 0.7946 & & 0.8768 &0.8216 &&0.8860&0.7969 \\   
\textbf{AUC} &  & 0.9725 & 0.9280& & 0.9709 &0.9216 &&0.9688&0.9324\\ 
\bottomrule
	\end{tabular}
	\label{A1}
\end{table}

\begin{figure}[!htb]
	\subfigure[Training Performance Comparison for $k=10$]
	{ \includegraphics[width=0.52\linewidth, height=0.40\linewidth]{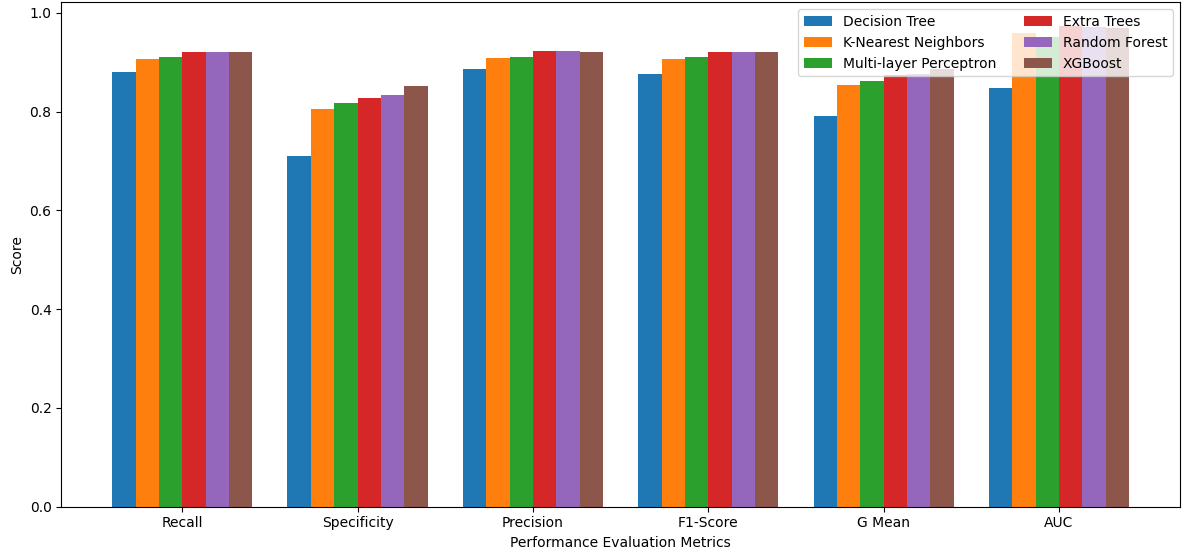}
		\label{k10a}
	}\hfill
	\subfigure[Validation Performance Comparison for $k=10$]
	{ \includegraphics[width=0.52\linewidth,height=0.40\linewidth]{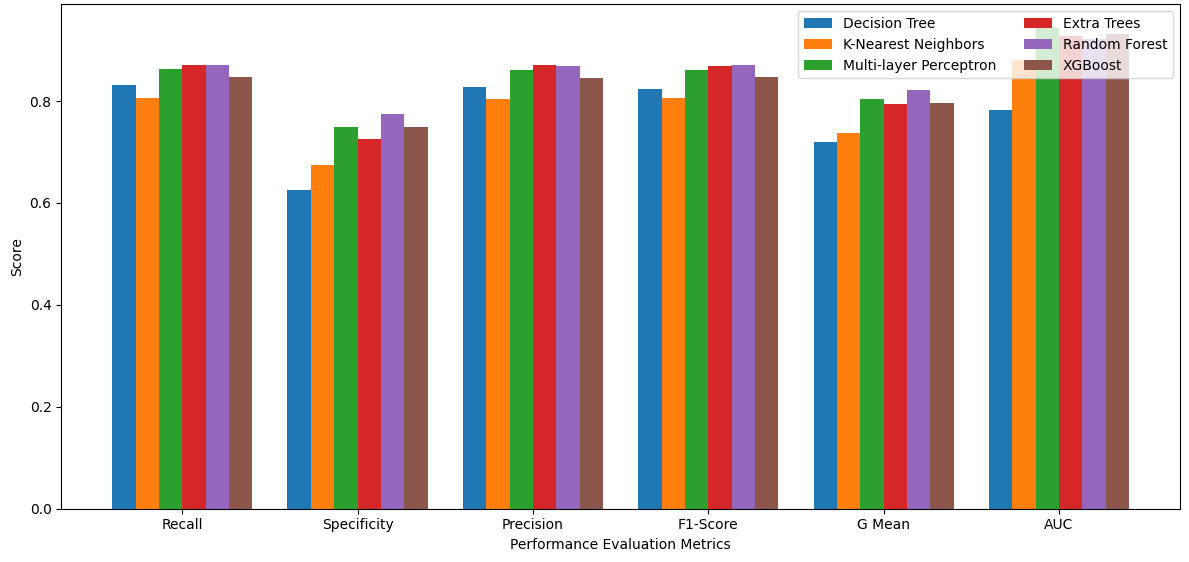}
		\label{k10b}
	}
	\caption{\textbf{Training and Validation Performance Comparison for $k=10$}}
	\label{k10}
\end{figure}

\noindent

\newpage
\subsection{Insightful Findings for $k=15$}
Rigorous hyparameter tuning via GridSearchCV resulted in the following optimal hyperparameters  for the six (6) models for $k=15$. Decision Trees: criterion=\texttt{gini}, max\_depth=3, max\_features=161, min\_samples\_leaf=2, min\_samples\_split=2.  Random Forest: criterion=\texttt{entropy}, max\_depth=None, max\_features=\texttt{sqrt}, min\_samples\_leaf=2, min\_samples\_split=2, n\_estimators=100.  Extra Trees: criterion=\texttt{gini}, max\_depth=10, max\_features=\texttt{sqrt}, min\_samples\_leaf=2, min\_samples\_split=5, n\_estimators=100.  XGBoost: colsample\_bytree=1.0, gamma=1, learning\_rate=0.1, max\_depth=5, n\_estimators=100, subsample=0.6. K-Nearest Neighbors: metric=\texttt{euclidean}, n\_neighbors=5, weights=\texttt{uniform}. Multi-Layer Perceptron: activation=\texttt{relu}, alpha=0.0001, hidden\_layer\_sizes=(100,), learning\_rate=\texttt{constant}, solver=\texttt{adam}.

\begin{table}[!htbp]
	\centering
	\caption{Model Performance Evaluation of DT, KNN, MLP, NB, RF and XGBoost for $k=15$}
	\begin{tabular}{lccccccccc}
		\toprule
		\textbf{Models}	& &  \multicolumn{2}{c}{\textbf{Decision Tree}} && \multicolumn{2}{c}{\textbf{K-Nearest Neighbors}}&&\multicolumn{2}{c}{\textbf{Multi-layer Perceptron}}\\ 
		
		\cmidrule{3-4} \cmidrule{6-7}\cmidrule{9-10}  \vspace{0.05in}
		&& Training   & Validation &&Training  & Validation  &&Training&Validation\\ 
		\hline
		\textbf{Recall (TPR)} 	&  & 0.8810 & 0.8306 & & 0.9173 &0.8387 &&0.9093&0.8790\\
		\textbf{Specificity (TNR)} &  & 0.7101 & 0.6250 & & 0.8225 &0.6750 &&0.8047&0.8000 \\ 
		\textbf{Precision} 	&  & 0.8855 & 0.8283 & & 0.9179 &0.8357 &&0.9099&0.8783 \\ 
		\textbf{F1-Scores} 	&  & 0.8766 & 0.8242 & & 0.9160 & 0.8350 &&0.9077&0.8786 \\ 
		\textbf{G-Mean} &  & 0.7909 & 0.7205 & & 0.8686 &0.7524 &&0.8554&0.8386 \\   
		\textbf{AUC} &  & 0.8480 & 0.7908 & & 0.9721 &0.8859 &&0.9543&0.9537 \\ 
		\bottomrule
		\textbf{Models}	& &  \multicolumn{2}{c}{\textbf{Extra Trees}} && \multicolumn{2}{c}{\textbf{Random Forest}}&&\multicolumn{2}{c}{\textbf{XGBoost}}\\ 
		
		\cmidrule{3-4} \cmidrule{6-7}\cmidrule{9-10}  \vspace{0.05in}
		&& Training   & Validation &&Training  & Validation  &&Training&Validation\\ 
		\hline
		\textbf{Recall (TPR)} 	&  & 0.9254 & 0.8790 & & 0.9415 &0.8790 &&0.9335&0.8871 \\
		\textbf{Specificity (TNR)} &  & 0.8225 & 0.7000 & & 0.8757 &0.7500 &&0.8639&0.8000 \\ 
		\textbf{Precision} 	&  & 0.9272 & 0.8814 & & 0.9418 &0.8778 &&0.9336&0.8861 \\ 
		\textbf{F1-Scores} &  & 0.9239 & 0.8744 & & 0.9409 &0.8767 &&0.9328&0.8863 \\ 
		\textbf{G-Mean} &  & 0.8724 & 0.7844 & & 0.9080 &0.8120 &&0.8980&0.8424 \\   
		\textbf{AUC} &  & 0.9706 & 0.9394 & & 0.9844 & 0.9379 &&0.9746&0.9461 \\ 
		\bottomrule
	\end{tabular}
	\label{A2}
\end{table}

\begin{figure}[!htb]
	\subfigure[Training Performance Comparison for $k=15$]
	{ \includegraphics[width=0.52\linewidth, height=0.40\linewidth]{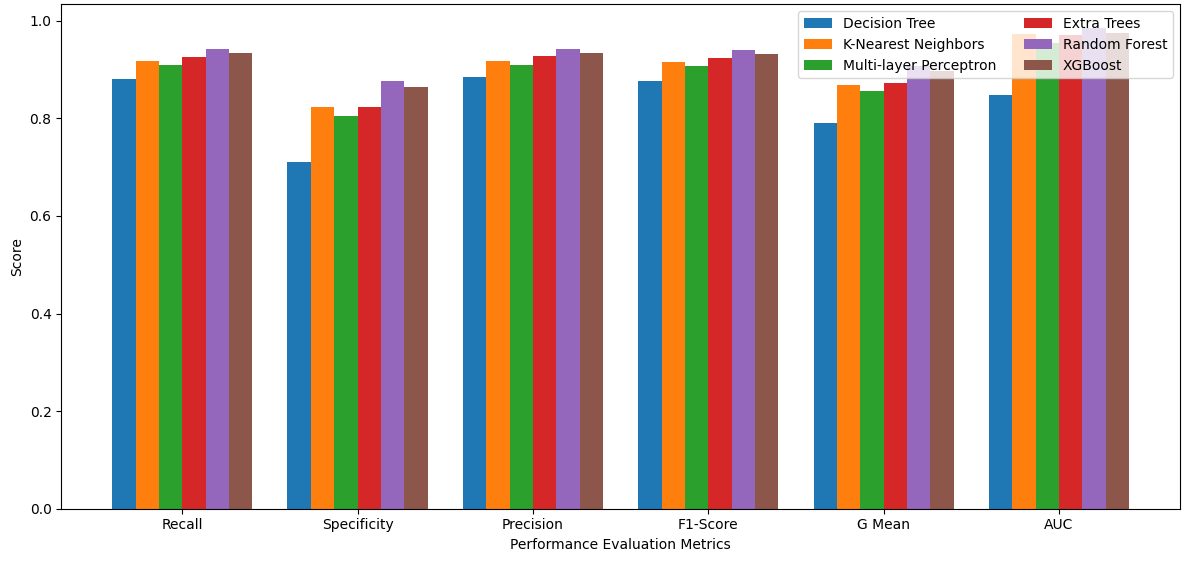}
		\label{k15a}
	}\hfill
	\subfigure[Validation Performance Comparison for $k=15$]
	{ \includegraphics[width=0.52\linewidth,height=0.40\linewidth]{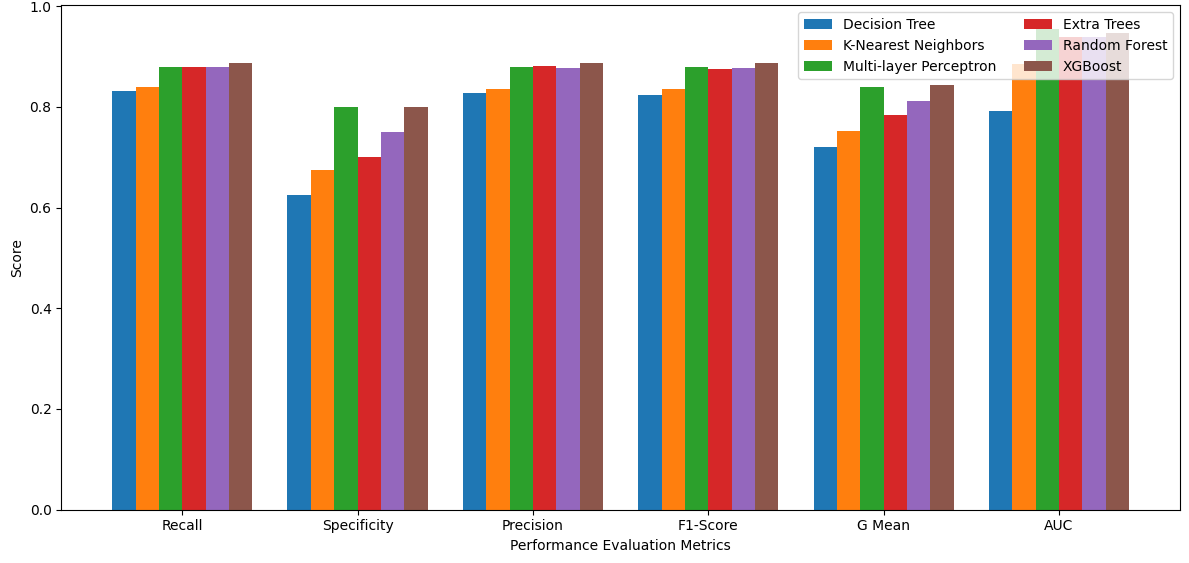}
		\label{k15b}
	}
	\caption{\textbf{Training and Validation Performance Comparison for $k=15$}}
	\label{k15}
\end{figure}

\newpage
\subsection{Insightful Findings for $k=20$}
Rigorous hyparameter tuning via GridSearchCV resulted in the following optimal hyperparameters  for the six (6) models for $k=20$. Decision Trees: criterion=\texttt{gini}, max\_depth=3, max\_features=161, min\_samples\_leaf=2, min\_samples\_split=2.  Random Forest: criterion=\texttt{entropy}, max\_depth=None, max\_features=\texttt{sqrt}, min\_samples\_leaf=1, min\_samples\_split=10, n\_estimators=200. Extra Trees: criterion=\texttt{gini}, max\_depth=None, max\_features=\texttt{sqrt}, min\_samples\_leaf=2, min\_samples\_split=5, n\_estimators=100. XGBoost: colsample\_bytree=1.0, gamma=1, learning\_rate=0.1, max\_depth=5, n\_estimators=50, subsample=0.6. K-Nearest Neighbors: metric=\texttt{euclidean}, n\_neighbors=3, weights=\texttt{uniform}.  Multi-Layer Perceptron: activation=\texttt{tanh}, alpha=0.0001, hidden\_layer\_sizes=(150,), learning\_rate=\texttt{constant}, solver=\texttt{adam}.

\begin{table}[!htbp]
	\centering
	\caption{Model Performance Evaluation of DT, KNN, MLP, NB, RF and XGBoost for $k=20$}
	\begin{tabular}{lccccccccc}
		\toprule
		\textbf{Models}	& &  \multicolumn{2}{c}{\textbf{Decision Tree}} && \multicolumn{2}{c}{\textbf{K-Nearest Neighbors}}&&\multicolumn{2}{c}{\textbf{Multi-layer Perceptron}}\\ 
		
		\cmidrule{3-4} \cmidrule{6-7}\cmidrule{9-10}  \vspace{0.05in}
		&& Training   & Validation &&Training  & Validation  &&Training&Validation\\ 
		\hline
		\textbf{Recall (TPR)} 	&  & 0.8810 &  0.8306 & & 0.9395 &0.8548 &&0.9274&0.9113 \\
		\textbf{Specificity (TNR)} &  & 0.7101 & 0.6250 & & 0.8521 &0.6750 &&0.8402&0.8000 \\ 
		\textbf{Precision} 	&  &  0.8855 &  0.8283 & &0.9411 &0.8539 &&0.9281&0.9115 \\ 
		\textbf{F1-Scores} 	&  &  0.8766 & 0.8242 & & 0.9385 & 0.8501 && 0.9264&0.9096 \\ 
		\textbf{G-Mean} &  & 0.7909 & 0.7205 & &  0.8947 &0.7596 &&0.8828&0.8538 \\   
		\textbf{AUC} &  & 0.8483 & 0.7911 & & 0.9863 & 0.8762 &&0.9654&0.9625 \\ 
		\bottomrule
		\textbf{Models}	& &  \multicolumn{2}{c}{\textbf{Extra Trees}} && \multicolumn{2}{c}{\textbf{Random Forest}}&&\multicolumn{2}{c}{\textbf{XGBoost}}\\ 
		
		\cmidrule{3-4} \cmidrule{6-7}\cmidrule{9-10}  \vspace{0.05in}
		&& Training   & Validation &&Training  & Validation  &&Training&Validation\\ 
		\hline
		\textbf{Recall (TPR)} 	&  & 0.9476 & 0.8710 & & 0.9577 &0.9577 &&0.9375&0.8468 \\
		\textbf{Specificity (TNR)} &  & 0.8876 & 0.8876 & & 0.9349 &0.7750 &&0.8817&0.7250\\ 
		\textbf{Precision} 	&  & 0.9479 & 0.8693 & & 0.9576 &0.8776 && 0.9373& 0.9373 \\ 
		\textbf{F1-Scores} &  & 0.9471 & 0.9471 & & 0.9576 & 0.8777 && 0.9371& 0.9371 \\ 
		\textbf{G-Mean} &  & 0.9171 & 0.8082 & &  0.8777 &0.8254 &&0.9091& 0.7835 \\   
		\textbf{AUC} &  & 0.9908 & 0.9446 & & 0.9958 &0.9280 &&  0.9822&0.9560 \\ 
		\bottomrule
	\end{tabular}
	\label{A3}
\end{table}

\begin{figure}[!htb]
	\subfigure[Training Performance Comparison for $k=20$]
	{ \includegraphics[width=0.52\linewidth, height=0.40\linewidth]{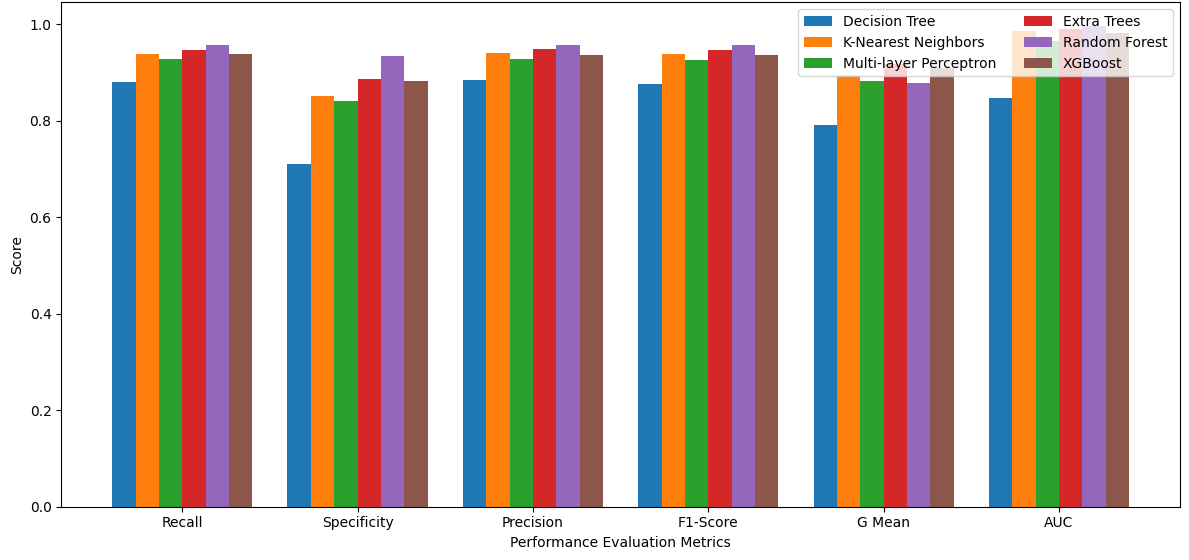}
		\label{k20a}
	}\hfill
	\subfigure[Validation Performance Comparison for $k=20$]
	{ \includegraphics[width=0.52\linewidth,height=0.40\linewidth]{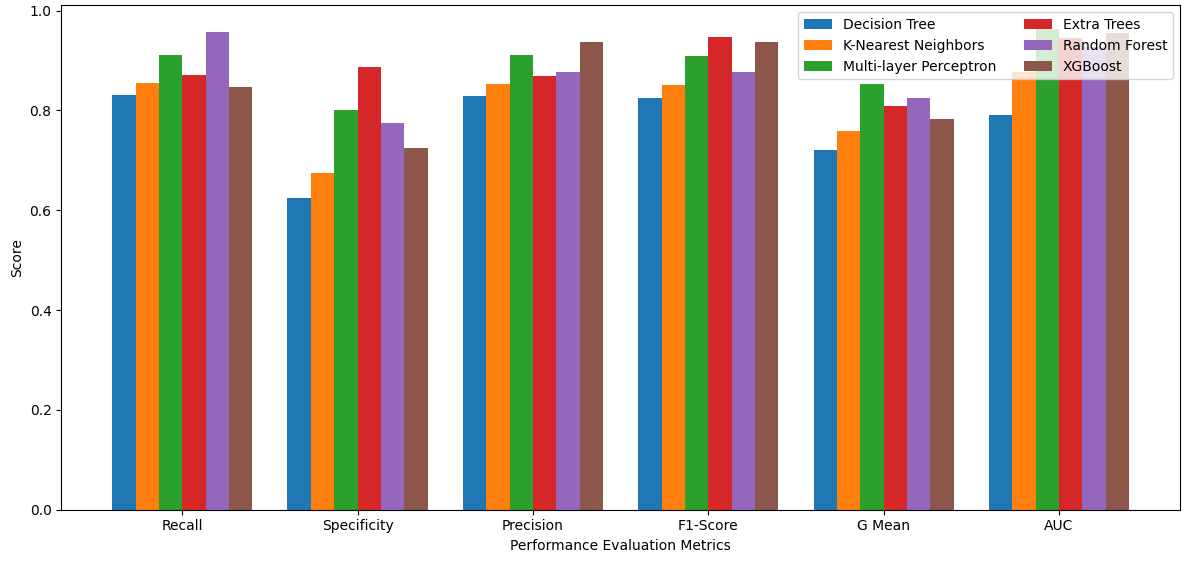}
		\label{k20b}
	}
	\caption{\textbf{Training and Validation Performance Comparison for $k=20$}}
	\label{k20}
\end{figure}

\newpage
\subsection{Insightful Findings for $k=25$}
Rigorous hyparameter tuning via GridSearchCV resulted in the following optimal hyperparameters  for the six (6) models for $k=25$. Decision Tree: criterion=\texttt{gini}, max\_depth=3, max\_features=161,min\_samples\_leaf=2, min\_samples\_split=2; Random Forest:
criterion=\texttt{gini}, max\_depth=10, max\_features=log2, min\_samples\_leaf=1, min\_samples\_split=2, n\_estimators=100; Extra Trees: criterion=\texttt{gini}, max\_depth=None, max\_features=log2,
min\_samples\_leaf=2, min\_samples\_split=2, n\_estimators=200; XGBoost: 
colsample\_bytree=1.0, gamma=5,\\
 learning\_rate=0.2, max\_depth=5, n\_estimators=100, subsample=0.6.
K-Nearest Neighbors: metric=euclidean, n\_neighbors=5, weights=\texttt{uniform};
Multi-Layer Perceptron: 
activation=\texttt{relu}, alpha=0.0001, hidden\_layer\_sizes=(150,), learning\_rate=constant, solver=adam.

\begin{table}[!htbp]
	\centering
	\caption{Model Performance Evaluation of DT, KNN, MLP, NB, RF and XGBoost for $k=25$}
	\begin{tabular}{lccccccccc}
		\toprule
		\textbf{Models}	& &  \multicolumn{2}{c}{\textbf{Decision Tree}} && \multicolumn{2}{c}{\textbf{K-Nearest Neighbors}}&&\multicolumn{2}{c}{\textbf{Multi-layer Perceptron}}\\ 
		
		\cmidrule{3-4} \cmidrule{6-7}\cmidrule{9-10}  \vspace{0.05in}
	\textbf{Metrics}	&& Training   & Validation &&Training  & Validation  &&Training&Validation\\ 
		\hline
		\textbf{Recall (TPR)} 	&  & 0.8810 & 0.8306 & & 0.9032 &0.8710 &&0.9274& 0.9113 \\
		\textbf{Specificity (TNR)} &  & 0.7101 & 0.6250 & & 0.7396 &0.6750 &&0.8521& 0.8250 \\ 
		\textbf{Precision} 	&  & 0.8855 & 0.8283 & & 0.9104 &0.8741 &&0.9275&0.9106 \\ 
		\textbf{F1-Scores} 	&  & 0.8766 & 0.8242 & & 0.8995 & 0.8654 &&0.9266&0.9103 \\ 
		\textbf{G-Mean} &  & 0.7909 & 0.7205 & & 0.8174 &0.7667 &&0.8889&0.8671 \\   
		\textbf{AUC} &  & 0.8483 & 0.7911 & & 0.9801&0.9003 &&0.9691&0.9690 \\ 
		\bottomrule
		\textbf{Models}	& &  \multicolumn{2}{c}{\textbf{Extra Trees}} && \multicolumn{2}{c}{\textbf{Random Forest}}&&\multicolumn{2}{c}{\textbf{XGBoost}}\\ 
		
		\cmidrule{3-4} \cmidrule{6-7}\cmidrule{9-10}  \vspace{0.05in}
	\textbf{Metrics}	&& Training   & Validation &&Training  & Validation  &&Training&Validation\\ 
		\hline
		\textbf{Recall (TPR)} 	&  & 0.9496 & 0.8871& & 0.9617 &0.8790 &&0.9254&0.8710 \\
		\textbf{Specificity (TNR)} &  & 0.8876 & 0.7250 & & 0.8876 &0.7000 &&0.8521&0.7250 \\ 
		\textbf{Precision} 	&  & 0.9501 & 0.8888 & &0.9638 &0.8814 &&0.9253& 0.8698 \\ 
		\textbf{F1-Scores} &  & 0.9491 & 0.8834 & & 0.9611 &0.8744 &&0.9246&0.8680 \\ 
		\textbf{G-Mean} &  &0.9181 & 0.8020 & & 0.9239 &0.7844 &&0.8880&0.7946 \\   
		\textbf{AUC} &  & 0.9908 & 0.9545 & & 0.9911 &0.9315 &&0.9698&0.9494 \\ 
		\bottomrule
	\end{tabular}
	\label{A4}
\end{table}

\begin{figure}[!htb]
	\subfigure[Training Performance Comparison for $k=25$]
	{ \includegraphics[width=0.52\linewidth, height=0.40\linewidth]{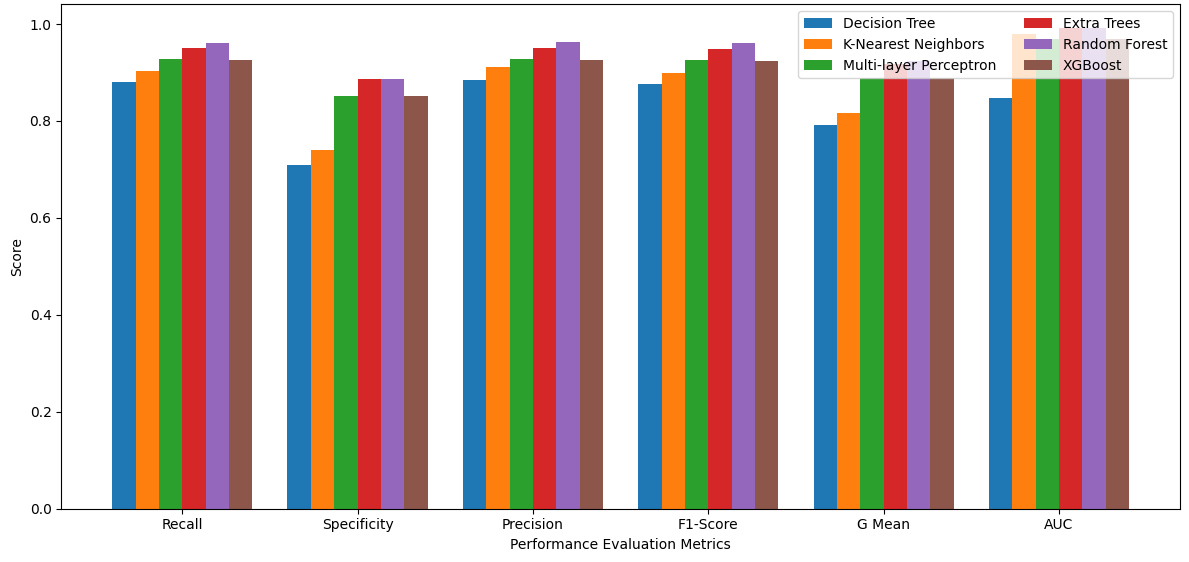}
		\label{k25a}
	}\hfill
	\subfigure[Validation Performance Comparison for $k=25$]
	{ \includegraphics[width=0.52\linewidth,height=0.40\linewidth]{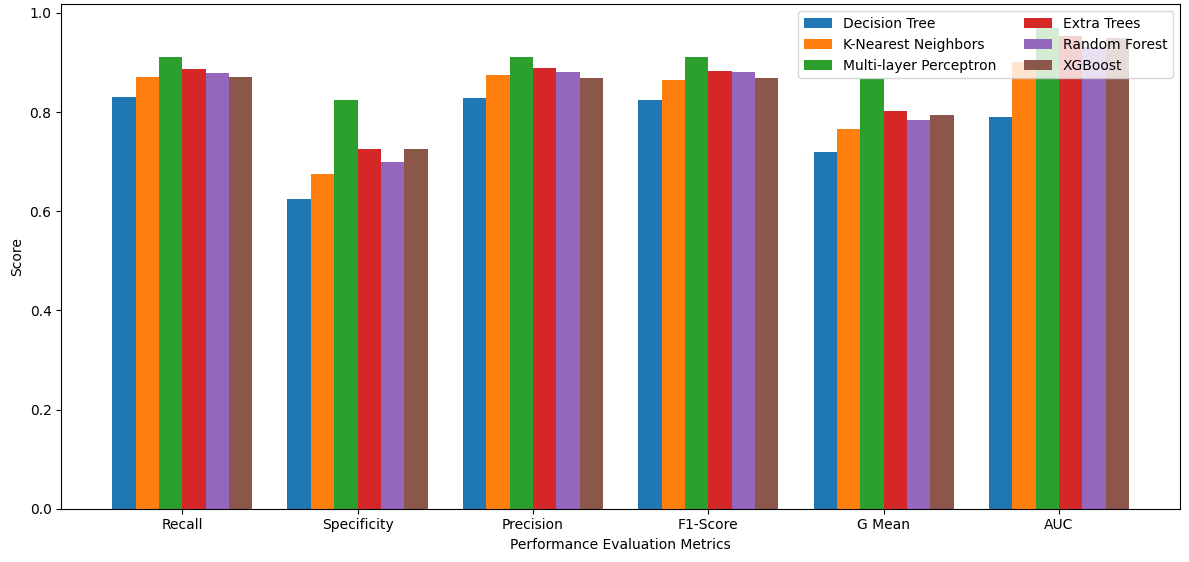}
		\label{k25b}
	}
	\caption{\textbf{Training and Validation Performance Comparison for $k=25$}}
	\label{k25}
\end{figure}

\newpage
\subsection{Discussion on Tables \ref{A1}-\ref{A4}}
Tables \ref{A1}-\ref{A4} present the performance of six machine learning models namely Decision Tree, K-Nearest Neighbors, Multi-layer Perceptron, Extra Trees, Random Forest, and XGBoost across six (6) key metrics. These metrics include Recall (TPR), Specificity (TNR), Precision, F1-Score, G-Mean, and AUC, with an emphasis on validation performance to assess each model’s ability to generalize beyond the training data.\\

\noindent
In Table \ref{A1}, for the feature set size $k=10$, the performance across models shows that Extra Trees and Random Forest clearly dominate in terms of validation metrics. Random Forest achieved a validation F1 score of 0.8701, G-mean of 0.8216, and AUC of 0.9216 while Extra Trees was very close with a achieved a validation F1 score of 0.8680, a G-mean of 0.7946, and an AUC of 0.9280. The MLP model showed solid performance with an F1 score of 0.8614 and a G-mean of 0.8045. XGBoost also performed well but had a slightly lower F1 score (0.8463) on validation, although its AUC was comparable at 0.9324. On the other hand, simpler models like K-Nearest Neighbors (KNN) and Decision Tree lagged behind, with the Decision Tree showing clear signs of overfitting given the larger gap between training and validation scores. Extra Trees and Random Forest strike a good balance between high predictive power and generalization.\\

\noindent
When the feature set was increased to $k=15$ in Table \ref{A2}, the performance of all models improved slightly. XGBoost emerged as a strong contender with the highest validation F1 score of 0.8863, complemented by a G-mean of 0.8424 and an AUC of 0.9461, suggesting robust classification ability. MLP improved with a validation F1 score of 0.8786, G-mean of 0.8386, and a slightly higher AUC of 0.9537.
Random Forest maintained impressive performance with a validation F1 score of 0.8767, G-mean of 0.8120, and a slightly higher AUC of 0.9379. Extra Trees followed closely behind with similar metrics. The MLP model also showed improvement with a validation F1 of 0.8786 and a G-mean of 0.8386. KNN gained traction but still lagged behind the ensemble methods. Decision Tree continued to underperform with a noticeable gap between training and validation, indicating persistent overfitting issues. This iteration reinforces the strength of ensemble models, particularly XGBoost and Random Forest together with MLP as the best performers for $k=15$.\\

\noindent
From Table \ref{A3}, at $k=20$, the results further emphasize the strength of the ensemble models. Extra Trees exhibited exceptional validation metrics, with an F1 score soaring to 0.9471 and a G-mean of 0.8082, alongside a high AUC of 0.9446. XGBoost also showed remarkable results with a validation F1 score of 0.9371 and an AUC of 0.9560, maintaining a strong balance between sensitivity and specificity. While Random Forest had an outstanding training AUC (nearly perfect), the validation AUC dipped to 0.9280, signaling mild overfitting. The MLP continued to perform robustly, showing high validation F1 score (0.9096) and good generalization. KNN improved but remained behind the top performers. The Decision Tree again showed the weakest validation results and signs of overfitting. Extra Trees, XGBoost and MLP clearly stand out at this feature set size for their strong validation performance with moderate overfitting risks.\\

\noindent
For the largest feature set tested with $k=25$ in Table \ref{A4}, Extra Trees achieved a validation F1 of 0.8834, G-mean of 0.8020, and AUC of 0.9545, signaling consistent and strong predictive power. The MLP model displayed high validation F1 (0.9103) and an even higher G-mean of 0.8671, indicating excellent balance and generalization. Random Forest remained competitive, with a validation F1 score of 0.8744 and a high AUC of 0.9315. XGBoost’s validation F1 dropped slightly to 0.8680 but retained a solid AUC of 0.9494. KNN and Decision Tree performed worse relative to the other models, with Decision Tree again showing the most overfitting concerns. This final comparison confirms that Extra Trees, MLP, and Random Forest are the top models, successfully balancing accuracy with generalization at this level of feature complexity.\\

\noindent
Across all four feature set sizes ($k=10,15,20,25$), MLP, Extra Trees, Random Forest, and XGBoost models consistently rank among the top performers for this coffee rating prediction task based on validation F1 score, G-mean, and AUC. Simpler models like KNN and Decision Tree consistently underperform and show overfitting issues.

\newpage
 \begin{figure}[!htb]
	\subfigure[Comparison of Feature Score by Class for $k=10$]
	{ \includegraphics[width=0.50\linewidth]{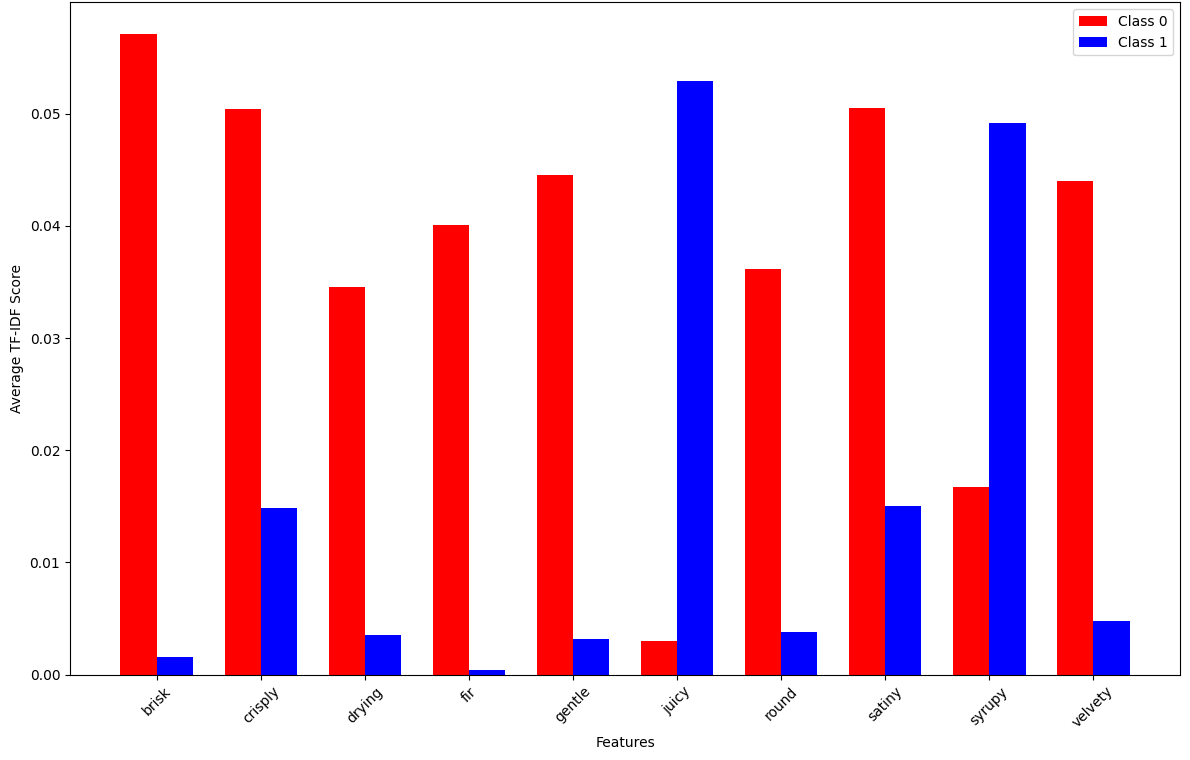}
		\label{S1}
	}\hfill
	\subfigure[Comparison of Feature Score by Class for $k=15$]
	{ \includegraphics[width=0.50\linewidth]{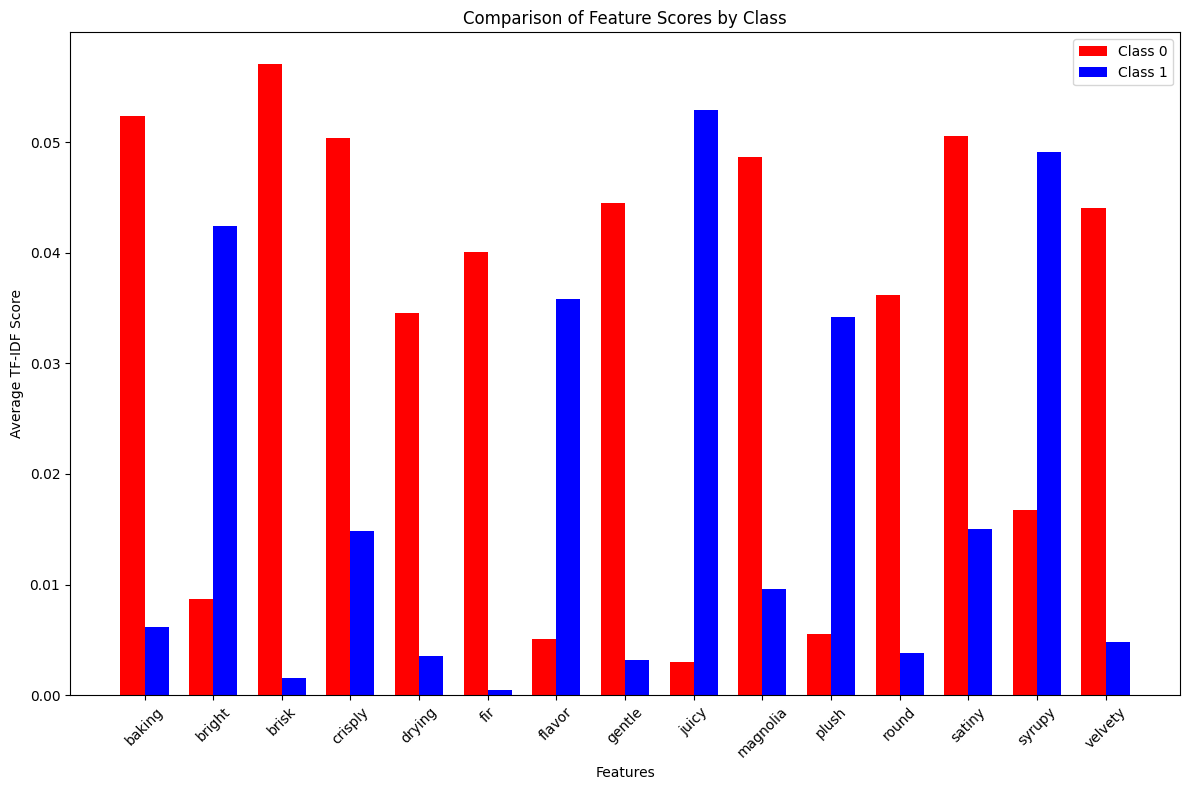}
		\label{G1}
	}\vfill
	\subfigure[Comparison of Feature Score by Class for $k=20$]
	{ \includegraphics[width=0.50\linewidth]{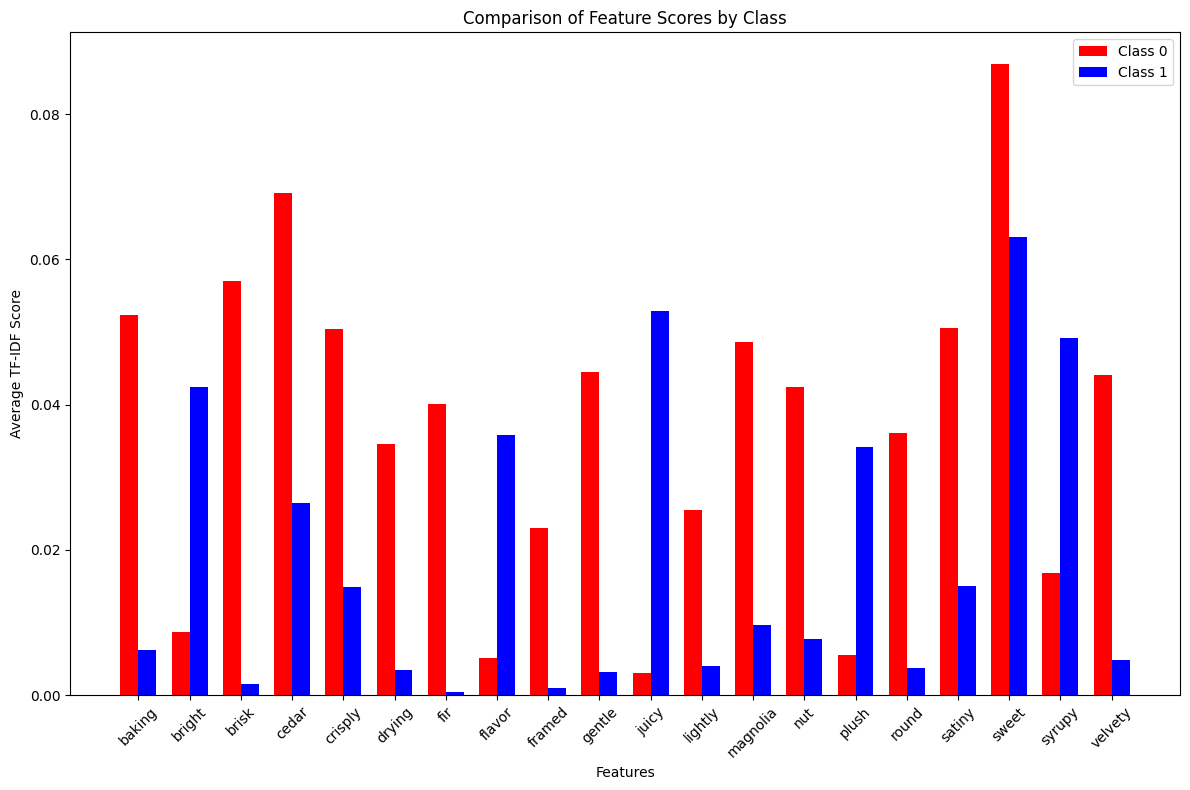}
	\label{H1}
	}\hfill
	\subfigure[Comparison of Feature Score by Class for $k=25$]
	{ \includegraphics[width=0.50\linewidth]{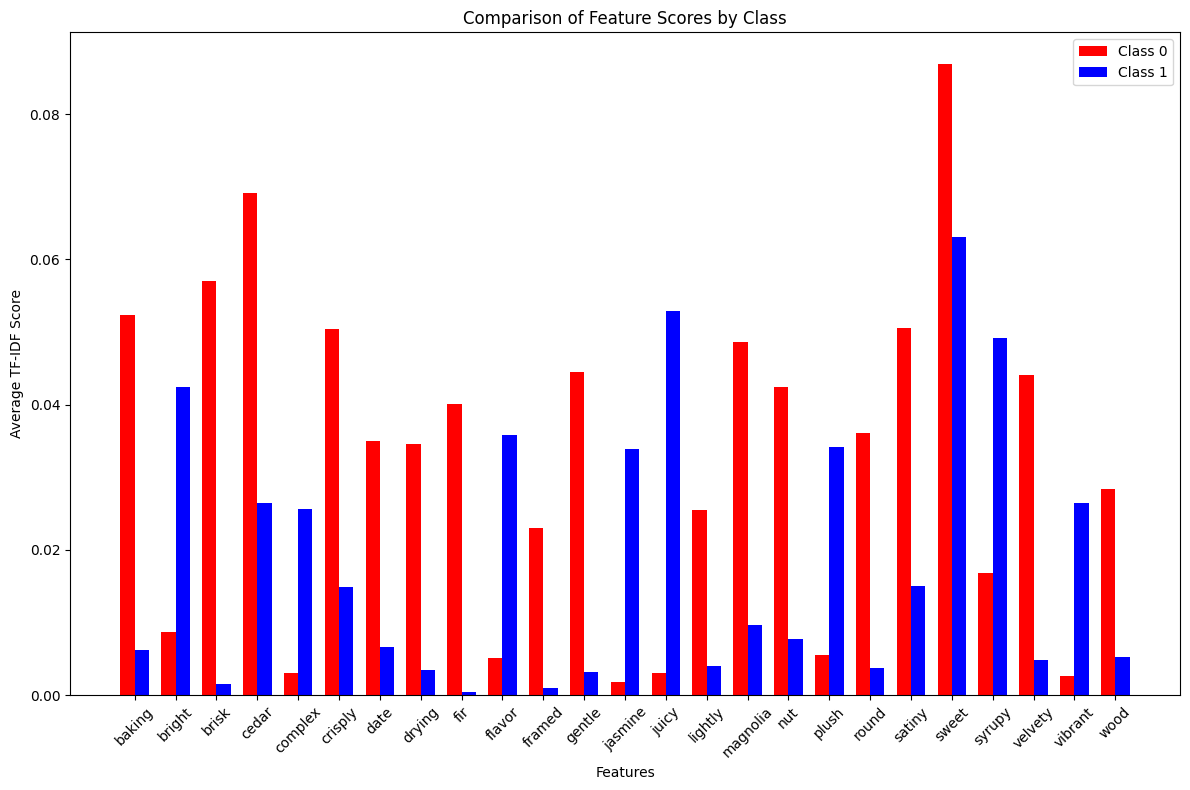}
	\label{I1}
	}
	\caption{\textbf{Feature scores comparison by class}}
	\label{Inf}
\end{figure}

\noindent
From Figure \ref{Inf}, \texttt{SelectKBest} was employed to select 10, 15, 20 and 25 features that contributed the most to distinguishing between coffee ratings where 0 means average coffee and 1 means
outstanding coffee. These features include brisk, crisply, drying, fir, gentle, juicy, round, satiny, syrupy, velvety, among others.

\subsection{Challenges Encountered}
Despite these promising outcomes, the project was not without its challenges. Some notable challenges encountered in the execution of this study are outlined below:

\begin{itemize}
	\item \textbf{Data Preprocessing:} Cleaning and transforming text data proved challenging due to the variability in reviews, requiring careful handling of stopwords, punctuation, and lemmatization to ensure meaningful feature extraction.
	\item \textbf{Overfitting:} Some models, particularly the Decision Tree, showed signs of overfitting when trained on a large number of features. Cross-validation and hyperparameter tuning helped mitigate this issue.
	\item \textbf{Feature Selection:} Identifying the most relevant features was difficult, as irrelevant features could negatively impact model performance. Techniques like \texttt{SelectKBest} was employed.
\end{itemize}

\newpage
\section{Conclusion and Recommendation \label{sec5}}
In this study, we set out to explore how supervised machine learning algorithms, when combined with robust feature selection methods, could be leveraged to predict coffee ratings based on influential textual and numerical attributes. By systematically preprocessing raw review data and extracting meaningful features through techniques like TF-IDF and SelectKBest, we were able to convert unstructured information into a form suitable for computational modeling. Our results demonstrate that ensemble methods such as Extra Trees, Random Forest, and XGBoost consistently outperform simpler algorithms like K-Nearest Neighbors and Decision Trees, particularly when it comes to balancing accuracy and generalization across various feature set sizes. Notably, the Multi-layer Perceptron (MLP) also showed strong performance, often matching or exceeding the ensemble models in terms of F1-score and G-mean, especially as more features were introduced.\\

\noindent
One of the central findings of this work is the critical role that careful feature selection and hyperparameter tuning play in boosting predictive performance. Models trained on optimally selected features (using SelectKBest) not only achieved higher validation metrics but also exhibited reduced tendencies toward overfitting, a challenge that was particularly apparent with the decision tree model. These insights reinforce the importance of both model complexity and data representation in developing reliable predictive systems within the domain of sensory product evaluation. This study illustrates the transformative potential of machine learning in automating and refining the assessment of coffee quality. Data-driven methods are expected to become more and more significant in helping producers, consumers, and researchers both as they develop, so promoting a more open, consistent, and insightful assessment process for one of the most loved beverages.\\

\noindent
Future research on predicting coffee quality should focus on addressing current challenges by using larger and more varied datasets that include different coffee types, origins, and processing methods. This extension of the dataset will enhance model predictability, improve generalization to unprocessed data, and provide a more robust investigation of deep learning applications in the coffee business. Future research should also aim to improve coffee classification approaches by including advanced structures like transformers or hybrid models, which may improve performance in differentiating visually identical coffee varieties.
\subsection*{Declaration of competing interest}
\noindent
The authors declare that they have no known competing financial interests or personal relationships that could have appeared to
influence the work reported in this paper.

\section*{Author Contributions}
All authors declare to have contributed equally to this project. All authors read and approved 
of the final manuscript submitted for publication. 

\subsection*{Data Availability}
\noindent
The data used to support the findings of this study is available upon reasonable request from the corresponding author.

\subsection*{Acknowledgment}
\noindent
The first and corresponding author  acknowledges the enormous support of the University of Tulane Dean Research  Council Scholarship and the University of Texas Rio Grande Valley (UTRGV) Presidential Research Fellowship Fund.

\renewcommand{\bibname}{References}
\bibliographystyle{unsrt}
\bibliography{REFERENCES}

\end{document}